# Open Knowledge Base Canonicalization with Multi-task Learning

Bingchen Liu[1], Huang Peng[2], Weixin Zeng[2*], Xiang Zhao[2], Shijun Liu[1,3*], Li Pan[1]

[1] School of Software, Shandong University, 1500 Shunhua Road, Jinan, 250000, Shandong Province, China.
[2] Laboratory for Big Data and Decision, National University of Defense Technology, 109 Deya Road, Changsha, 410073, Hunan Province, China.
[3] QuanCheng Laboratory, Jinan, 250103, Shandong Province, China.

*Corresponding author(s). E-mail(s): zengweixin13@nudt.edu.cn; lsj@sdu.edu.cn;

**Abstract**

The construction of large open knowledge bases (OKBs) is integral to many knowledge-driven applications on the worldwide web such as web search. However, noun phrases and relational phrases in OKBs often suffer from redundancy and ambiguity, which calls for the investigation on OKB canonicalization. Current solutions address OKB canonicalization by devising advanced clustering algorithms and using knowledge graph embedding (KGE) to further facilitate the canonicalization process. Nevertheless, these works fail to fully exploit the synergy between clustering and KGE learning, and the methods designed for these sub-tasks are sub-optimal. To this end, we put forward a multi-task learning framework, namely MulCanon, to tackle OKB canonicalization. In addition, diffusion model is used in the soft clustering process to improve the noun phrase representations with neighboring information, which can lead to more accurate representations. MulCanon unifies the learning objectives of these sub-tasks, and adopts a two-stage multi-task learning paradigm for training. A thorough experimental study on popular OKB canonicalization benchmarks validates that MulCanon can achieve competitive canonicalization results.

# 1 Introduction

With the progress in closed information extraction (CIE) techniques [1], curated knowledge bases (CKBs) like YAGO [2] and Freebase [3] have developed rapidly. CKBs make a profound impact on various knowledge-driven applications on the worldwide web such as web search [4] and news recommendation [5]. However, new facts continuously emerge in the real world, which may not be covered by existing ontology in CKBs, rendering CKBs inadequate to handle time-sensitive tasks.

As an alternative, open information extraction (OIE) is put forward [6], which extracts triples in the form of ⟨*head noun phrase*, *relational phrase*, *tail noun phrase*⟩ from the unstructured text; in this scenario, there is no need to designate an ontology in advance. The extracted triples together constitute a large open knowledge base (OKB), such as ReVerb [7]. While OKB comes with advantages over CKB in terms of coverage and diversity, there is a prominent issue associated with OKB, i.e., noun phrases in the triplets are not canonicalized, which may lead to redundancy and ambiguity of knowledge facts [8]. This can further affect downstream tasks. For instance, in the field of news recommendation [5], the redundancy and ambiguity in expression can affect the effectiveness and user experience.

In response, the task of OKB canonicalization is intensively studied and plays a significant role [9], where synonymous noun phrases are aggregated into the same cluster, and thus the OIE triples can be converted into canonicalized forms. An example is provided in Example 1 and Figure 1. This greatly eliminates the redundant information and resolves the ambiguity of knowledge facts, providing a refined knowledge base for the downstream tasks.

**Example 1.** *As shown in Figure 1, there are three sentences concerning Messi extracted from the news, where an OIE system extracts three OIE triples. Unfortunately, it is not clear whether Messi and Lionel Messi refer to the same entity. Hence, when the term Lionel Messi is queried, the machine may not effectively present all available facts related to the entity. In addition, it can be seen that the first two OIE triples refer to the same meaning, which can be merged and reduce the redundancy. The canonical results are provided in the right of Figure 1.*

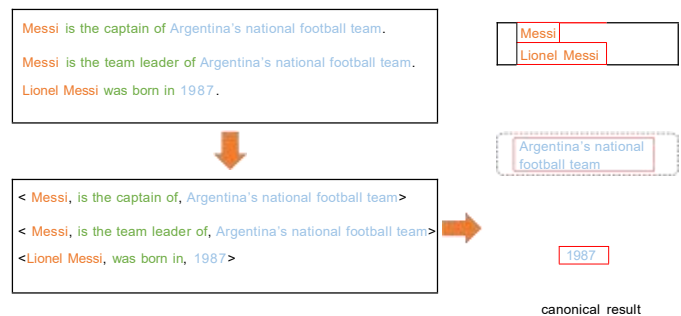

**Fig. 1** An example of OIE and canonical results.

In existing literature, while some OKB canonicalization approaches learn the knowledge graph embedding (KGE) of OIE triples to conduct the clustering [10], some use the information in the source context to facilitate the clustering process [11]. Some fuse the two types of information sources in pursuit of a more accurate clustering result [12]. A recent work adopts variational autoencoder (VAE), a generative method, to tackle OKB canonicalization, which learns the embeddings and cluster assignments in an end-to-end fashion, resulting in better vector representations for noun phrases, and also competitive canonicalization results [13].

Nevertheless, existing studies fail to exploit the mutual interactions among the sub-tasks in canonicalization, such as clustering and KGE learning, resulting in less satisfying canonicalization performance. In addition, they obtain the representations of noun phrases based on the textual features, with the multi-hop neighboring information overlooked.

To address the above issues, we propose MulCanon, a novel framework for open knowledge base canonicalization with *two-stage multi-task learning*. Specifically, as the main purpose is to assign noun phrases with the same meaning into the same cluster, we use a generative framework, variational deep embedding (VaDE) [14], to conduct the soft clustering, where the original VAE model is replaced by diffusion model to avoid the potential information loss during the transformation, which is referred to as the *diffusion sub-task*. Meanwhile, we use hierarchical agglomerative clustering (HAC) upon the phrases to obtain an initial cluster assignment, which is contrasted with the clustering result produced by the diffusion model, constituting the *clustering sub-task*. In addition, the *KGE sub-task* is conducted to ensure that the canonicalized noun phrases satisfy the inherent KG structure. Finally, following [8], we also consider the *side information sub-task* to use more comprehensive signals for canonization. Notably, we enhance the noun phrase representation with the neighboring information to facilitate the clustering process.

We integrate the aforementioned sub-tasks and use a ***two-stage multi-task learning*** paradigm for model training, which can better capture the interactions among sub-tasks, thus regulating the canonicalization process and leading to better results. Considering that training KGE requires the outputs of clustering, in stage 1, we eliminate the KGE sub-task to obtain relatively stable parameters of the clustering module. Then in stage 2, we integrate all sub-tasks to train a more comprehensive model, yielding more accurate clustering results. Extensive experimental results validate that our proposal can consistently achieve superior canonization results than state-of-the-art solutions on standard benchmark.

The major contributions of this work are as follows:

- We put forward MulCanon, a novel framework for open knowledge base canonicalization. We integrate separated learning objectives and use a two-stage multi-task paradigm for model training.
- We use the diffusion model in the soft clustering process, and improve the noun phrases representations with neighboring information.
- Extensive experimental results demonstrate that MulCanon can achieve competitive canonicalization results on standard benchmarks.

In Section 2, we present the work related to OKB canonicalization. In Section 3, we specify our proposed model and approach. In Section 4, we conduct experiments to verify the effectiveness of the proposed approach. Finally, we conclude the paper in Section 5.

## 2 Related work

In this section, we present the relevant works related to OKB canonicalization.

**OKB canonicalization.** Research on OKB canonicalization can be divided into two types: semi-supervised methods [15][16][8][17] and unsupervised methods [11][12]. Regarding the unsupervised methods, Lin et al. [11] use relevant knowledge from the context of the extracted source text to aggregate noun phrases into groups with similar respective interpretations. In order to compensate for the shortcomings of using knowledge from only a single perspective, CMVC [12] use knowledge from both views to obtain more accurate results.

As to the semi-supervised methods, the work [15] applies HAC algorithms to manually defined features such as overlapping IDF [15] tokens to assign synonymous noun phrases into the same group. Wu et al. [17] use pruning and boundary techniques to reduce the similarity computation and build a graph-based clustering approach based on the previous canonicalization model [15] , so as to improve the efficiency and applicability of the model. CESI [8] first learns the embedding of noun phrases using a KGE model based on the factual view and additionally using various edge information (e.g. PPDB information [18], etc.), and performs the embedding based on the embedding to bring the interpretation of the same noun phrases into a group. JOCL [16] explores a clustering scheme combining a canonicalization task and a linking task using a factor graph model. Unlike state-of-the-art approaches that focus on equating the OKB canonicalization problem to the clustering problem of noun phrases, ignoring the generation of more accurate representations of noun phrases, CUVA [13] applies variational autoencoders (VAE) to the process of learning noun phrases representations.

**Knowledge graph alignment.** The purpose of knowledge graph alignment is to determine whether two or more entities from different sources are pointing to the same object in the real world, and to bring together named entities that have the same referent object in different knowledge graphs [19–22]. For instance, to alleviate the limitation of relying only on structural information, Zeng et al. [23] introduced entity name information and designed a common attention feature fusion network to adjust the weights of different features to improve the performance of the knowledge graph entity alignment. This task is similar to OKB canonicalization, as they both aim to reduce the redundant entities and create a refined knowledge graph. Nevertheless, OKB canonicalization involves the canonicalization of elements in a single knowledge graph, while KG alignment tends to take place among multiple KGs.

**Multi-task learning.** Multi-task learning integrates the training process of different tasks with certain correlations, and then jointly trains the model parameters of several different tasks simultaneously to improve the learning ability of the model and enhance the generalization of the model. Multi-task learning has been shown to have superior performance in applications such as multi-text categorization [24], recommender systems [25] [26], and so on. In knowledge graph related tasks, multi-task learning has also been used for knowledge graph completion [27] and knowledge graph embedding [28]. In this paper, we use multi-task learning for the OKB canonicalization task, fully exploiting the use of associations in each sub-task for model training.

**Diffusion model.** The diffusion model defines a markov chain of diffusion steps to slowly add random noise to the data, and then learns the inverse diffusion process to construct the required data samples from the noise [29]. Over recent years, diffusion models have achieved significant advantages in many areas such as anomaly detection [29], image generation [30] and image noise removal [31]. In this work, we also use the diffusion model in the OKB canonicalization process, which can help to learn data representations and lead to better clustering results.

## 3 Methodology

In this section, we first define the problem, describe the framework outline of our proposed model, and present each part of the model separately.

### 3.1 Problem definition

Open Information Extraction (OIE) targets at open texts and aims to extract a large number of relational tuples without any relation-specific training data. The extracted triples are stored in the form of (noun phrases, relational phrases, noun phrases), forming a large Open Knowledge Base (OKB).

The goal of the OKB canonicalization task is to use the triples and the information extracted from the source text s to cluster synonymous noun phrases pointing to the same entity into a group, thus converting these OIE triples into the canonical forms.

### 3.2 Framework outline

As shown in Figure 2, the inputs include the OKB triples, consisting of head noun phrases (in green), relational phrases (in orange), and tail noun phrases (in blue), the contextual information of these triples, and pre-trained word embeddings such as GloVe [10]. We use Glove word embeddings here, in order to ensure consistency with the use of word embeddings across baseline models and thus ensure fairness in the comparison of experimental results. The goal is to generate the canonical forms of these phrases, as shown in the rightmost of Figure 2.

We devise a multi-task learning framework, MulCanon, to tackle OKB canonicalization. The main task is to divide the phrases into their corresponding cluster, and we use a generative framework, variational deep embedding (VaDE) [14], to conduct the soft clustering. Instead of adopting the original VAE model, we propose to utilize the diffusion model to avoid the potential information loss during the transformation, resulting in the diffusion loss $L_{diff}$. In the meantime, we use HAC [32] upon the phrases to obtain an initial cluster assignment, which is also used to initialize the Gaussian mixture model. Then, based on the Gaussian mixture model and the middle variable **w** of the diffusion process, we predict the cluster assignment, which is contrasted with the clustering result produced by HAC, leading to the clustering loss $L_{clu}$. Note that the results produced by HAC are merely considered as *weak supervision*, and there is no actual supervision signal. As an additional task, KGE is conducted to ensure that the canonicalized noun phrases satisfy the inherent KG structure. Specifically, we use the predicted cluster assignment to obtain the representations for elements in OIE triples, and use the HolE model [33] to produce the KGE loss $L_{kge}$. We also consider the side information loss $L_{side}$ by following previous work.

Finally, we formulate the multi-task learning objective L and train the model, which is then used to infer the final canonicalization results. The training is divided into two stages:

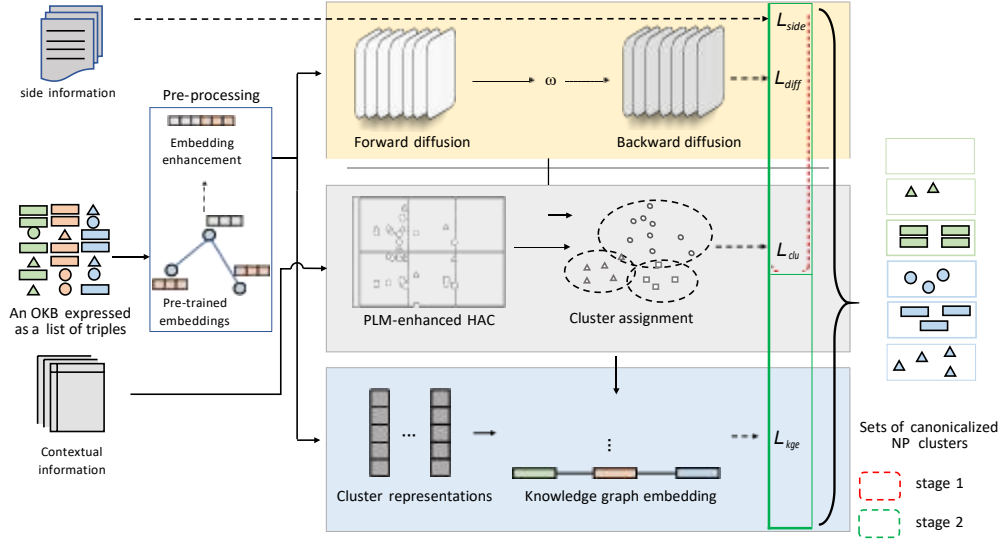

**Fig. 2** The main framework of MulCanon.

the first stage involves training the diffusion model, clustering process, and side information modeling; In the second stage, the KGE loss is also incorporated into the training process.

The main notations are summarized in Table 1.

### 3.3 Pre-processing

We consider that the initial representations of noun phrases directly obtained by using GloVe vectors cannot adequately represent their corresponding features. Thus, we propose to integrate the features of first-order neighboring entities to further enhance the representations. The first-order neighboring entities refer to the entities directly connected to the given entity in the triple relationship information stored in the OKB.

Formally, for each noun phrase information, its initial representation $\mathbf{e}_i$ is first generated based on GloVe [10]. For the i-th noun phrases, the augmented representation $\mathbf{h}_i$ is obtained by concatenating the means of its first-order neighbors:

$$\mathbf{h}_i = [\mathbf{e}_i \parallel \frac{1}{|\mathcal{N}_i|} \sum_{j \in \mathcal{N}_i} \mathbf{e}_j], \tag{1}$$

where $\mathcal{N}_i$ denotes the ordinal numbers of all first-order neighboring noun phrases of the i-noun phrase, and $[\cdot \parallel \cdot]$ denotes the vector splicing operation.

### 3.4 Cluster assignment

Following existing literature [12], we consider OKB canonicalization as a clustering problem, and thus the main goal is to correctly predict the cluster assignment.

**Table 1** Important notations

| Notation | Description |
|---|---|
| $e_i$ | initial representation of noun phrases |
| $h_i$ | augmented representation of noun phrases |
| $L_{diff}$ | diffusion loss |
| $L_{clu}$ | clustering loss |
| $L_{kge}$ | knowledge graph embedding loss |
| $L_{side}$ | side information objective |
| $p(c)$ | prior probability of cluster assignment |
| $\mu_c$ | mean of the mixed Gaussian distribution corresponding to cluster c |
| $\sigma_c^2$ | variance of the mixed Gaussian distribution corresponding to cluster c |
| $\beta_t$ | variance schedule across diffusion steps |
| t | number of steps |
| $z_1$ | noise sampled from the standard normal distribution |
| $x_{t-1}$ | representation in the previous step of diffusion process |
| T | step of diffusion process |
| $v(c)$ | final probability of assigning to cluster c |
| W | learnable matrix |
| z | noise that fits the standard normal distribution |
| $f_T(x_t, t)$ | neural network to approximate the conditional probabilities |
| $L_1$ | multi-task learning objective of stage one |
| $L_2$ | multi-task learning objective of stage two |

**PLM-enhanced HAC clustering algorithm.** We utilize the pre-trained language models (PLM) [34] and HAC clustering algorithm [35] to obtain a preliminary clustering result. PLM uses the contextual information as input to obtain the pre-trained embeddings. After that, preliminary clustering result of the noun phrases can be obtained from HAC clustering algorithm by taking the phrases and the pre-trained embeddings as input. The preliminary clustering results are selected for initialising the Gaussian mixture model. During this process, we also generate the preliminary labels for noun phrases.

**Clustering prediction.** As shown in the center of Figure 2, in MulCanon, the probability of the actual cluster assignment is calculated via a softmax function, where the probability of assigning to cluster c is denoted as:

$$v(c) = \frac{p(c)p(\omega|c)}{\sum p(c)p(\omega|c)}, \qquad (2)$$

where $p(c)$ is the prior probability, calculated as the percentage of phrases belonging to cluster c in the Gaussian mixture model. $p(\omega|c)$ denotes the probability of sampling $\omega$ from $VI(\mu_c, \sigma_c^2)$, where $\mu_c$ and $\sigma_c^2$ are the mean and variance of the mixed Gaussian distribution corresponding to the cluster c. $\omega$ is obtained by the diffusion model process, which will be detailed in the next subsection. Then, we consider the label generated by the HAC algorithm as weak supervision, and contrast it with the predicted result v using the cross-entropy cost function, leading to the clustering loss $\mathcal{L}_{clu}$. Notably, the labels produced by the HAC algorithm are only weak supervision, and the potential errors will be corrected by the multi-task learning process.

### 3.5 Diffusion model

We adopt a generative framework, VaDE [14], to conduct the soft clustering, which generates $\omega$ and forwards to the clustering assignment process. Specifically, we propose to utilize the diffusion model that performs equidimensional transformation, instead of the original VAE model used in [14].

**Forward diffusion process.** The forward diffusion process is a continuous noise addition process, calculated as follows:

$$\mathbf{x}_t = \sqrt{\alpha_t}\mathbf{x}_{t-1} + \sqrt{1-\alpha_t}\mathbf{z}_1, \tag{3}$$

$$\alpha_t = 1 - \beta_t, \tag{4}$$

where $\beta_t$ represents the variance schedule across the diffusion steps [29]. $t$ is the number of steps, $\mathbf{z}_1$ is the noise sampled from the standard normal distribution, $\mathbf{x}_{t-1}$ is the representation in the previous step, and $\mathbf{x}_0$ in the first step is initialized with the GloVe embedding of the corresponding noun phrases.

After obtaining the representations in the last step T, $\mathbf{x}_T$, we use the reparametrization trick to sample $\omega$:

$$\omega = \mu + \sigma \odot \mathbf{z}_0, \tag{5}$$

where $\mu = \mathbf{x}_T \mathbf{W}_\mu$ and $\sigma = \mathbf{x}_T \mathbf{W}_\sigma$. $\mathbf{W}_\mu$ and $\mathbf{W}_\sigma$ are two learnable matrices, and $\mathbf{z}_0$ is a noise that fits the standard normal distribution.

**Reverse diffusion process.** Then we conduct the reverse diffusion, which progressively removes noise, starting with a random noise and gradually reducing to a representation without noise:

$$\mathbf{x}_{t-1} = \frac{1}{\sqrt{\alpha_t}}\mathbf{x}_t - \frac{\sqrt{1-\alpha_t}}{\sqrt{\alpha_t}}f_\tau(\mathbf{x}_t, t) + \mathbf{z}_2, \tag{6}$$

where $\alpha_t$ is calculated using Equation 4, $\mathbf{z}_2$ is also a noise that fits the standard normal distribution, $f_\tau(\mathbf{x}_t, t)$ is a neural network to approximate the conditional probabilities, which is implemented by multilayer perceptron, and $\tau$ refers to the trainable parameter in the network.

**Diffusion loss.** The diffusion process aims to train $f_\tau$ so that its predicted noise is similar to the real noise used for destruction. Following existing literature [29], the diffusion loss $\mathcal{L}_{diff}$ is:

$$\mathcal{L}_{diff} = \mathbb{E}_{t\sim[0-T], f\sim\mathcal{N}(0,\mathbf{I})}[||f - f_\tau(\mathbf{x}_t, t)||^2], \tag{7}$$

where $f$ is sampled from a standard Gaussian distribution, $t$ denotes the time step, and $||\cdot||$ denotes the L2 distance.

Compared with the hard clustering strategy used in the majority of existing works, the soft clustering of the generative model can help to explain different meanings of a given entity mention. This advantage reasonably compensates for the shortcomings of hard clustering methods that only assign each entity to a cluster. That being said, traditional generative models such as generative adversarial network (GAN) and VAE involve dimensional compression and expansion, which might lead to information distortion and loss during data transformation. Hence, in this work, we fill in this gap by using the diffusion model.

### 3.6 Knowledge graph embedding

Referring to the initial embedding representation generated based on Glove [10] and the preliminary clustering allocation results based on Section 3.4. We consider the corresponding embedding with the highest predicted clustering probability among them as the input to the KGE. MulCanon produce the KGE loss $L_{kge}$ based on knowledge graph embedding and use the HolE model [33] as the calculation method here:

$$L_{kge} = \text{HolE}(h, r, t), \qquad (8)$$

where HolE represents the KGE loss mentioned in [33], and h,r,t represent the the corresponding embedding with the highest predicted clustering probability on Section 3.4.

### 3.7 Side information

Following exsiting works [8], we also adopt the side information to aid the canonicalization, including:

- PPDB information: PPDB 2.0 [36] is used here to determine whether two noun phrases are equivalent to each other. This is done by extracting high-confidence paraphrases from the dataset and removing overlapping redundancies in the first step. In the second step, the information extracted in the previous step is found in a concatenated set, all equivalent noun phrases are clustered, and a representation is defined for each cluster. In the third step, a search is performed to find out whether two noun phrases have the same representation and thus determine the equivalence between them.
- Entity linking: compared to the process of PPDB information, this information is relatively simple to obtain. To obtain entity linking information, it is only necessary to use the entity linker [37] for a given text to map the noun phrases to the entities present in Wikipedia. The equivalence between them is determined based on whether they are linked to the same entity or not.
- Morphological canonicalization: find equivalent noun phrases using the morphological canonicalization method used in open information extraction [38].
- IDF token overlap: if noun phrases share infrequently used terms, they are considered to have a higher probability of referring to the same entity, which is considered a valid feature for use. For the specific formula for calculating overlap scores, we refer to the formula provided in [8], and we remove pairs with scores less than a specific threshold.

We adopt the side information objective $L_{side}$ by following [13] and [8]. The main motivation is to harness the contextual information to further facilitate the overall learning process.

### 3.8 Multi-task learning objective

As to the overall training procedure, we adopt a two-stage strategy. In stage one, the multi-task learning objective is:

$$L_1 = L_{clu} + L_{diff} + L_{side}, \tag{9}$$

In stage two, the multi-task learning objective is:

$$L_2 = L_{clu} + L_{diff} + L_{kge} + L_{side}, \tag{10}$$

Noteworthily, by conducting the multi-task learning, MulCanon can better capture the interactions among sub-tasks, which can further regulate the canonicalization process and lead to better results. Note that the KGE loss $L_{kge}$ is obtained by the TransE model [39]. The motivation for adopting the two-stage training strategy is that training KGE requires the outputs of clustering. Thus, by training the generative clustering process first, we can obtain relatively stable parameters of the clustering module, based on which the KGE can be better trained.

---

**Algorithm 1** The main process of Mulcanon

**Inputs:**
OKB=(h,r, t): an OKB expressed as a list of triples
$S_n$: the contextual information of the above triples
$S_{side}$: the side information

**Output:**
$Set_1, \ldots, Set_n$: Sets of canonicalized clusters

1: Generate the initial representation $e_i$ with the Glove embedding using OKB
2: Generate the augmented representation $h_i$ using Eq.(1) based on $e_i$
3: Generate a preliminary clustering results of OKB and preliminary labels by PLM-enhanced HAC clustering algorithm based on $S_n$
4: Calculate $x_t$ using Eq.(3)-Eq.(4) using $h_i$ as $x_0$
5: Calculate $\omega$ using Eq.(5) using $x_t$
6: Calculate the probability of assigning to cluster c to obtain sets of canonicalized clusters $Set_1, \ldots, Set_n$ using Eq.(2)
7: Calculate $x_{t-1}$ using Eq.(6)
8: Using Eq.(7) to obtain the diffusion loss $L_{diff}$
9: Following [13] and [8] to obtain the side information objective $L_{side}$ based on $S_{side}$
10: Using the cross-entropy cost function to obtain the clustering loss $L_{clu}$
11: Using Eq.(8) to obtain the learning objective of stage one $L_1$ and optimize the model
12: Using the HolE model [33] to obtain the KGE loss $L_{kge}$
13: Using Eq.(9) to obtain the learning objective of stage two $L_2$ and optimize the model
14: return $Set_1, \ldots, Set_n$

---

## 4 Experiments

In this section, we conduct experiments to verify the effectiveness of our proposed model.

## 4.1 Experimental setting

**Implementation details.** As introduced in Section 3.8, we adopt a two-stage training paradigm. In the first stage, we mainly train the clustering, side information and diffusion models, and do not include KGE training. The learning rate is set to $10^{-3}$. In the second stage, we train KGE and all training content for the first stage. The learning rate is set to $10^{-5}$. The number of training rounds for each batch is set to 50. In the input side, if an noun phrase contains two or more words, its embedding is defined as the average of the embeddings of the individual words contained in it. As to the diffusion model, we set the time step T to 2, which will be further analyzed in the hyper-parameter analysis. For the loss function associated with the KGE algorithm, we randomly select 20 negative samples for each positive sample. In addition, experiments are conducted using one Intel x86 CPU and one NVIDIA GeForce RTX 3090 GPU with a maximum of 24GB RAM, implemented in Python 3.9.0.

**Evaluation metrics.** We adopt the F1 metric for evaluating entity canonicalization, which is the harmonic mean of precision (P) and recall (R). Following previous works [8] [15], we report results of three variants, i.e., macro F1, micro F1 and pair F1. Specifically, the introduction of various indicators is as follows:

- Macro. P is defined as the fraction of pure clusters in C, that is, all noun phrases are linked to clusters of the same gold entity. The calculation of R is similar to that of P, but the functions of E and C are interchangeable.
- Micro. P is defined as cluster C, which is based on the assumption that the most frequent gold entities in the cluster are correct. The definition of R is similar to the macro R.
- Pair. P is measured by the ratio of the number of hits in C to the total possible matches in C. R is the ratio of hits in C to all possible pairsin E. If a pair of elements in the cluster in C all refer to the same gold entity, a hit will occur.

More details are omitted in the interest of space, which can be found in [8]. Noteworthily, the average F1 is used as the primary evaluation metric.

**Datasets.** Previous works use the ReVerb45K [8] dataset for evaluation. The OIE triples of ReVerb45K are extracted by ReVerb [7] from the source text of Clueweb09 [40], and the noun phrases are annotated with the corresponding entities in Freebase. The number of triples in ReVerb45K is 45K, each triple of which is associated with a Freebase entity. For each entity, there are two or more aliases in the noun phrases.

Nevertheless, recent works [12,41] points out that the contextual information in ReVerb45K is incomplete, and even some labels are wrong. Hence, Jiang et al. [41] puts forward COMBO, a new benchmark extracted from the large Ontological KG Wiki-data, which has 18K OIE triples. This is a new related dataset that can be used for news augmentation [42] by matching news articles with Wiki-data. Compared to the existing dataset, this new dataset additionally provides gold ontology-level canonicalization of noun phrases, and source sentences from which triples are extracted. This dataset contains 18K triples and their source sentences, and more relevant auxiliary annotations are provided. In comparison with other existing datasets, this dataset has the longest average triple length and the largest number of unique noun phrases.

Therefore, to ensure fair comparison, we mainly compare the baseline methods on COMBO. We also provide the results on ReVerb45K as supplementary results. 20 percent of the triples are used for validation, and the rest of triples are regarded as the test set. The validation set is used to tune the hyper-parameters and there are no labeled data.

**Table 2** Performance of the noun phrase canonicalization task in COMBO.

| Model | Macro F1 | Micro F1 | Pair F1 | Average F1 |
|---|---|---|---|---|
| random+HAC [41] | 0.759 | 0.819 | 0.515 | 0.698 |
| Glove+HAC [41] | 0.889 | 0.899 | 0.612 | 0.800 |
| Glove+HolE+HAC [8] | 0.874 | 0.874 | 0.509 | 0.752 |
| Glove+SI+HAC [8] | 0.887 | 0.896 | 0.595 | 0.793 |
| CESI [8] | 0.889 | 0.899 | 0.564 | 0.784 |
| VAEGMM+SI [13] | 0.888 | 0.897 | 0.605 | 0.797 |
| VAEGMM+HolE [13] | 0.886 | 0.896 | 0.603 | 0.795 |
| CUVA [13] | 0.889 | 0.898 | 0.606 | 0.798 |
| Bert-base [41] | 0.914 | 0.925 | 0.751 | 0.864 |
| MulCanon | 0.915 | 0.927 | 0.768 | **0.870** |

**Methods for comparison.** Regarding the methods to compare, we follow COMBO [41] and compare with a set of baselines, which are detailed below:

- CESI [8], which fuses the knowledge embeddings of OIE triples and side information in pursuit of a more accurate clustering result. "SI" means the side information. "random" and "Glove" correspond to each other, which both belongs to the initialization embedding method, "random" means randomly initializing the initial embedding representation of noun phrases, and "Glove" means using GloVe [10] for initialization, "HAC " means hierarchical agglomerative clustering [32], "HolE" means the use of HolE embedding [43]. CESI first initializes the embedding by the initialization embedding method, then optimizes the embedding based on HolE embedding, and finally performs clustering operations based on HAC algorithm on the optimized embedding.
- CUVA [13], which introduces the generative model VAE into the OKB canonicalization and considers both relational information in the fact triples and side information. "VAEGMM" means the use of a Mixture of Gaussians within the latent space of a variational autoencoder [44]. CUVA adopts VAEGMM to jointly learn and cluster the embedding by the initialization embedding method and optimizes that by side information.
- Bert-base [41] is one of the approaches for data pre-processing based on pre-trained language models (PLM), by virtue of BERT's linguistic representation model [34] for the initial solution of clustering of trained models. It is also one of the baselines established by COMBO [41].

Notably, the performance of some baseline models, such as JOCL [16] and CMVC [12], are not provided in both main experiment and supplementary experiment, since they require additional pre-processing files. Hence, we follow COMBO and do not compare with them. Table 2 reports the noun phrase canonicalization results on COMBO.

## 4.2 Main results

From Table 2, it can be seen that the average F1 of our proposed MulCanon model is higher than the state-of-the-art work CUVA. CESI fully exploits the side information to facilitate the canonicalization process, which greatly improves the accuracy of clustering. The employment of side information has also been exploited by most of the recent methods. CUVA introduces the generative model VAE into the OKB canonicalization and considers both relational information in the fact triples and side information, ensuring a competitive result in the task of OKB canonicalization. In Table 2, we compare the experimental results for some variants of the CESI and CUVA methods.

Table 3 Performance of the noun phrase canonicalization task in ReVerb45K.

| Model | Macro F1 | Micro F1 | Pair F1 | Average F1 |
| --- | --- | --- | --- | --- |
| CESI | 0.640 | 0.855 | 0.842 | 0.779 |
| CUVA[1] | 0.601 | 0.846 | 0.830 | 0.759 |
| MulCanon | 0.751 | 0.833 | 0.795 | **0.793** |

[1] Notably, we try our best to re-implement the model using the open-source codes, while we cannot obtain the performance reported in the original paper [12].

Different from all baseline models, MulCanon adopts a multi-task learning framework, which can fully exploit the synergy among clustering, diffusion, KGE learning and side information modeling to tackle the canonicalization task. Particularly, in the learning of the representation of vectors, the splicing of first-order neighboring entities is used to make full use of the neighborhood information to supplement and enhance the corresponding representation. For these reasons, MulCanon attains quite competitive performance so far.

**Supplementary experiment.** In the supplementary experiments, we report the results on the ReVerb45K dataset. Neighborhood enhancement is not used here because the contextual information in the ReVerb45K dataset is not as sufficient as in the COMBO dataset, making it difficult to capture enough first-order neighborhood information. The results of the supplementary experiments are presented in Table 3, which verify that MulCanon is competitive compared to some current baseline models in the ReVerb45K dataset.

Table 4 Ablation study on the COMBO dataset.

| | Macro F1 | Micro F1 | Pair F1 | Average F1 |
| --- | --- | --- | --- | --- |
| MulCanon | 0.915 | 0.927 | 0.768 | **0.870** |
| w/o neighbor | 0.855 | 0.872 | 0.665 | 0.797 |
| w/o diffusion | 0.409 | 0.170 | 0.001 | 0.193 |
| w/o SI | 0.914 | 0.925 | 0.751 | 0.863 |
| w/o HolE | 0.914 | 0.927 | 0.761 | 0.867 |

## 4.3 Ablation study

We further conduct ablation study on COMBO to test the effectiveness of the multi-task learning and embedding enhancement in MulCanon. Specifically, we compare with w/o neighbor, which removes the embedding enhancement, w/o SI, which removes the side information modeling objective, w/o diffusion, which removes the diffusion model objective. w/o glove, which removes initialization using Glove, w/o random, which removes random initialization, w/o HolE, which represents replacing HolE [33] with TransE [39].

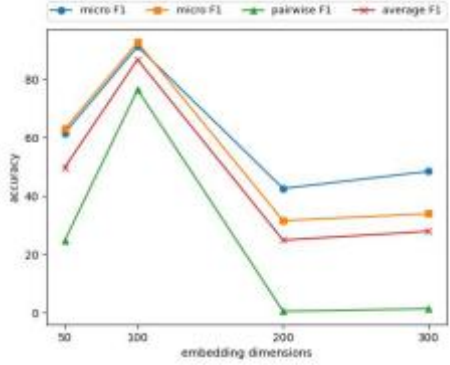 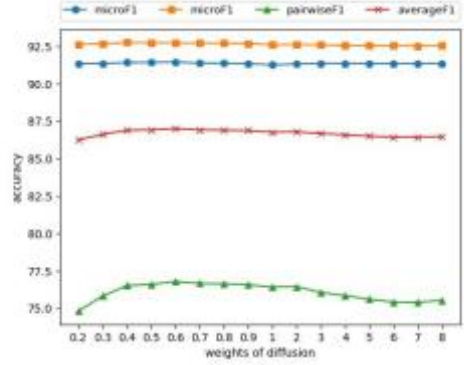

**Fig. 3** Hyper-parameter analysis on the embedding dimensions.

**Fig. 4** Hyper-parameter analysis on the weights of diffusion model.

As can be observed from Table 4, removing the diffusion model can cause the most significant performance drop, validating its importance. Besides, removing w/o neighbor also brings notable performance drop, thus validating for the importance of using domain information to enhance the corresponding input representation. Additionally, it shows in Table 4 that removing the objectives applied to subtasks also decreases the performance. These demonstrate that the novel mechanisms that we introduce into the canonicalization task are crucial to its superior performance. In all, by fully exploiting the connections among the subtasks and the method of initializing embeddings, our multi-task learning strategy can lead to competitive canonicalization results. It can also see that using different knowledge graph embedding models has little impact on the overall performance of the model.

### 4.4 Hyper-parameter analysis

**On the embedding dimensions.** To analyze the influence of the embedding dimensions on the effectiveness of MulCanon, we conduct the parameter analysis. Specifically, we set the the embedding dimensions to 50, 100, 200 and 300, respectively, and report the average F1 results in Figure 3. It observes that the embedding dimensions do affect the results, and results reach its maximum value at a value of 100, because it is not possible to accurately represent the features of noun phrases when the dimensionality is too small, and it is difficult to maintain convergence during training when the dimensionality is too large. Therefore, we set the value of embedding dimensions to 100 to allow the model to be optimised. Therefore, we set the value of embedding dimensions to 100 to allow the model to be optimised.

**On the weights of diffusion model.** We further investigate the influence of the weights of diffusion model on the overall results. We vary the weights of the diffusion model from 0.2 to 8 for multiple coefficient sizes to see the effect of the diffusion model weights on the overall effect of the model. We can observe that when the weights of the diffusion model change from 0.2 to 0.6, the overall trend of the evaluation indicators shows an upward trend and reaches the highest value at 0.6, and after 0.6, the overall trend of the evaluation indicators shows a downward trend. This shows that the overall performance of the model in the dataset is highest when the parameters of the diffusion model are at 0.6, so we set the weight of the final diffusion model in the model to 0.6. The corresponding *macro F1*, *micro F1*, *pair F1* and *average F1* results are reported in Figure 4.

It can be seen that, overall speaking, varying the weights do not have significant influence on the results, and all losses in the multi-task learning objective contribute positively to the overall results.

### 4.5 Case study for neighborhood information enhancement

Due to the textual similarity and the fact that the previous methods only generated representations based on the textual names of the noun phrases, these noun phrases are easily and wrongly considered to be the same noun phrases, and thus will be grouped into a cluster for the wrong canonicalization classification process. We give some such examples in Table 5. In detail, for example, taking **san juan** and **san jose** as an example, after adding the information of first-order neighboring noun phrases, **san juan**'s neighboring noun phrases are **hiram bithorn stadium** and **isla verde international airport**, and **san jose**'s neighbors are **cinequest film festival** and **solopower**. Although the texts of **san juan** and **san jose** are more similar, their first-order neighbor noun phrases are more different and thus will not be misclassified as the same entity. And when the neighbor information is removed, these similar phrases tend to be attributed to the same grouping again and thus misclassified.

**Table 5** Error cases of not using neighborhood information enhancement.

| Examples | Noun phrase | Noun phrase |
|----------|-------------|-------------|
| group 1  | london      | london midland |
| group 2  | cao gan     | cao cao     |
| group 3  | san juan    | san jose    |
| group 4  | street art  | street artists |
| group 5  | cool girl   | girl        |

## 5 Conclusion and future work

This paper proposes a multi-task learning framework (MulCanon) for open knowledge base canonicalization, which is of great significance to various knowledge-driven applications on the world wide web such as web search. Extensive experimental results on standard benchmarks demonstrate that our proposed model outperforms some existing models on the OKB canonicalization task. In future work, we will explore combining more subtasks to better complete the tasks of OKB canonicalization. In addition, we will further explore how to reduce the impact of expression ambiguity and redundancy on responsible and trustworthy news recommendations.

## Declarations

Ethical approval This declaration is not applicable.
Competing interests I declare that all authors have no competing interests as defined by Springer, or other interests that might be perceived to influence the results and discussion reported in this paper.

**Authors' contributions** Bingchen Liu wrote the main manuscript text, prepared all figures and tables and provided the methodology. Weixin Zeng, Xiang Zhao and Huang Peng provided writing-review and editing. Li Pan and Shijun Liu provided writing-review and editing and provided funding support.

**Funding** The authors would like to acknowledge the support provided by the National Key R&D Program of China under Grant 2023YFC3304904, the Shandong Provincial Natural Science Foundation of China under Grant ZR2023LZH016, the "New 20 Regulations for Universities" funding program of Jinan (202228089), and the TaiShan Industrial Experts Programme (tscx202312128).

**Availability of data and materials** All of the materials including figures is owned by the authors and no permissions are required.